\documentclass[runningheads]{llncs}
\usepackage[T1]{fontenc}
\usepackage{graphicx}
\usepackage{verbatim}
\usepackage{amsmath}
\usepackage{multirow}
\usepackage{placeins}
\usepackage{algorithm}
\usepackage{algpseudocode}
\usepackage{booktabs}
\usepackage{subcaption}
\usepackage{afterpage}
\usepackage[hyphens]{url} 

\begin{document}
\title{DINO-Med: A Unified Patch-Based Adaptation Framework for Multi-Modal Medical Image Analysis Applied to Liver Fibrosis Staging}
\titlerunning{DINO-Med: Patch-Based Adaptation for Liver Fibrosis}

\author{Boya Wang\inst{1}\thanks{Corresponding author: Boya.WANG@nottingham.ac.uk}\and Ruizhe Li\inst{1,2} \and Chao Chen\inst{1} \and Xin Chen\inst{1}} 
\authorrunning{Boya Wang et al.}

\institute{School of Computer Science, University of Nottingham, UK \and 
Nottingham Biomedical Research Centre (BRC),\\School of Medicine, University of Nottingham, UK}
\maketitle              
\begin{abstract}
 Adapting natural-image foundation models like DINOv3 to multi-modal medical imaging is challenging due to the significant domain gap between natural color images and multi-channel medical scans. We present a unified, patch-based framework that processes raw multi-modal imaging through training-free registration, automated localization, and mask-filtered patch extraction. This architecture culminates in a hierarchical strategy that aggregates patch-level insights into subject-level diagnostics. Using liver fibrosis staging as a case study, we evaluate four patch-level feature representations: handcrafted Radiomics features, learned ResNet features, pre-trained foundation model SAM-Med2D features, and frozen DINOv3 features. To ensure a controlled comparison, all models utilize the same lightweight MLP head and are evaluated across both rigid and deformable registration settings. Our training protocol focuses on mild fibrosis (S1) and cirrhosis (S4) classes only, enabling a single classifier to address both substantial fibrosis detection and cirrhosis staging. Evaluated via 10 random train (90\%)/ test (10\%) splits on 360 subjects from the CARE 2025 Liver Track 4 cohort, our DINOv3-based framework significantly outperforms all baselines, achieving the best classification accuracy of 78.4\% for S1 and 75.8\% for S4.
 
\end{abstract}
\section{Introduction}
In the clinical domain, the ability to extract precise, high-dimensional features from medical imaging is fundamental to accurate diagnosis and personalized treatment planning. Unlike natural images, medical data, such as multi-parametric MRI, contains subtle textural patterns and complex anatomical relationships that are often imperceptible to the human eye. Effective feature learning and extraction are therefore essential to transform these raw signals into discriminative biomarkers that can differentiate disease stages with high sensitivity, forming the backbone of automated clinical decision support.

The recent emergence of large-scale, pre-trained foundation models has initiated a new era in medical image analysis. Driven by extensive datasets and self-supervised learning (SSL), these models act as robust, task-agnostic feature extractors. Segmentation-centric models such as Segment Anything Model (SAM) \cite{kirillov2023segment} and its medical domain-adapted variants, SAM-Med2D \cite{cheng2023sam} and SAM-Med3D \cite{wang2025sam}, have demonstrated exceptional zero-shot capabilities in mapping complex anatomical structures. Conversely, representation-centric models such as the DINO series (DINOv1 \cite{caron2021emerging}, DINOv2 \cite{oquab2023dinov2}, and the latest DINOv3 \cite{simeoni2025dinov3}) have set new standards in visual understanding. Trained on extensive natural image corpora, the DINO architecture generates highly discriminative dense representations that transfer exceptionally well to downstream tasks, often eliminating the need for exhaustive end-to-end fine-tuning \cite{caron2021emerging,oquab2023dinov2,simeoni2025dinov3}.

Despite the remarkable capabilities of vision foundation models for feature extraction, a significant "domain gap" persists between natural color images and multi-modality medical imaging. This discrepancy in intensity distributions, spatial resolution, and dimensionality often limits the direct applicability of these models in clinical settings. Consequently, determining the optimal strategy to adapt pre-trained encoders and effectively fuse multi-modal representation features remains a critical and worthwhile challenge to explore. 

In this work, we present a generic, unified patch-based adaptation framework designed to fully unlock the representational power of the DINOv3 foundation model for multi-modality medical imaging. Our approach does not merely propose a high-performing task-specific pipeline, but fundamentally investigates how natural-image foundation models can be effectively adapted to handle the complexities of multi-modality medical data for subject-level classification tasks.

To rigorously test the clinical relevance and robustness of our generic framework, we apply it to the challenging task of liver fibrosis staging. Early detection and accurate staging of liver fibrosis are crucial for effective clinical management and patient prognosis, as untreated fibrosis may deteriorate into cirrhosis or hepatocellular carcinoma. Although percutaneous liver biopsy currently serves as the diagnostic standard, its routine clinical translation is fundamentally limited by procedural invasiveness, susceptibility to sampling errors, and the risk of severe complications \cite{bedossa2003sampling,bravo2001liver}. To overcome these limitations, non-invasive techniques, specifically multi-parametric magnetic resonance images (MRIs), have rapidly become the primary option for assessing fibrotic disease \cite{banerjee2014multiparametric,loomba2020advances}. By applying our framework to this specific task, we demonstrate that an optimally adapted DINOv3 feature extractor can effectively address these clinical challenges, providing a robust, non-invasive alternative for accurate disease staging. The main contributions of this paper are summarized as follows:

\begin{enumerate}
  \item A Unified Patch-Based Adaptation Framework: We propose a comprehensive, end-to-end diagnostic pipeline capable of processing raw, unaligned multi-modal medical images. The workflow integrates training-free registration \cite{wang2026searchmindtrainingfreemultimodalmedical}, automated localization, and mask-guided patch extraction, culminating in a hierarchical classification strategy that aggregates patch-level insights into subject-level diagnostics. To ensure a rigorous, unbiased evaluation, all upstream preprocessing and downstream aggregation modules are strictly standardized across all experimental cohorts. 
  
  \item Frozen DINOv3 Multi-Modal Descriptors: We leverage the pre-trained DINOv3 backbone as a frozen feature extractor, creating a potent multi-modal descriptor through independent feature concatenation. This simple, fine-tuning-free approach establishes a robust performance baseline, demonstrating the model's capacity for high-fidelity medical representation learning.

  \item Clinical Benchmarking in Liver Fibrosis: Using liver fibrosis staging as a rigorous testbed based on the MICCAI CARE 2025 Liver Track 4 dataset \cite{liu2025merit}, we evaluate our framework against radiomics, task-specific ResNet \cite{wang2026semi}, medical domain fine-tuned SAM-Med2D  baselines. Our results show that frozen DINOv3 consistently outperforms all baselines significantly across rigid and deformable registration settings.

\end{enumerate}

\section{Related Works}

\subsection{Radiomics and Deep Learning for Feature Extraction}

In recent studies, deep learning and radiomics have played a central role in medical image analysis, improving diagnostic accuracy and clinical interpretability. Traditional radiomics approaches involve extracting high-dimensional features, such as shape, intensity, and texture features, from medical images to represent tissue characteristics \cite{gillies2016radiomics,park2019radiomics}. Deep learning architectures, in particular Convolutional Neural Networks (CNNs) \cite{targ2016resnet,krizhevsky2012imagenet,simonyan2014very}, have shown great potential for automatically extracting hierarchical representations from MRIs \cite{yasaka2018deep}. To capture broader contextual dependencies, Vision Transformers (ViTs) have recently been adopted, leveraging self-attention mechanisms to produce superior feature representations \cite{dai2021transmed,manzari2023medvit}. Current methodological trends emphasize the integration of multi-view learning and uncertainty modeling to improve robustness and interpretability simultaneously.

\subsection{Pretrained Foundation Models in Medical Imaging}

The emergence of large-scale, self-supervised visual foundation models has transformed representation learning in computer vision and is beginning to change the landscape of medical imaging analysis \cite{he2024foundation}. Foundation models have become increasingly important in medical image analysis due to their ability to produce transferable visual representations through large-scale pretraining, thereby reducing dependence on task-specific annotation. There are two main directions in recent works. One is adapting general-purpose vision backbones pre-trained on natural images. The DINO family \cite{caron2021emerging,oquab2023dinov2,simeoni2025dinov3} is the representative series that has shown strong transferability, high quality, and robust patch-level representations, and therefore has become a widely used backbone for downstream tasks. The other is directly developing for medical tasks by using large-scale clinical medical imaging datasets. In segmentation, SAM-Med2D \cite{cheng2023sam} and SAM-Med3D \cite{wang2025sam} both adapt the Segment Anything Model \cite{kirillov2023segment} to medical imaging datasets. For multi-modal learning, BiomedCLIP \cite{zhang2023biomedclip} has been trained on large-scale biomedical image-text datasets (PMC-15M) and can transfer performance strongly across diverse biomedical vision-language tasks. More recently, RadFM \cite{wu2025towards} has been extended to radiology. However, effective adaptation of natural-image foundation models to multi-modal medical imaging remains underexplored. 

\subsection{Liver Fibrosis Staging}

Liver fibrosis staging has transitioned from invasive biopsies toward non-invasive multi-parametric MRI (mpMRI), which offers superior soft-tissue contrast through complementary imaging sequences. Initial computational efforts primarily utilized radiomics, combining handcrafted features, such as shape, first-order intensity, and texture, with traditional classifiers or shallow neural networks \cite{park2019radiomics}. However, these methods were constrained by their heavy reliance on precise segmentation and manual feature engineering, limiting their scalability and robustness. With the maturation of deep learning, convolutional neural networks (CNNs) became the standard for direct representation learning \cite{yasaka2018liver}. More recently, patch-based approaches applied to mpMRI (e.g., Wang et al. \cite{wang2026semi}) have demonstrated strong performance on benchmarks like the MICCAI CARE2025 Liver Challenge \cite{liu2025merit}. Despite these gains, there remains a significant research gap in systematically exploring how state-of-the-art vision models, pretrained on natural images, can be effectively adapted to the application of multi-parametric MRI or multimodal medical imaging in general. 

\section{Methods}

\subsection{Overview}

\begin{figure}[!htb]
\includegraphics[width=\textwidth]{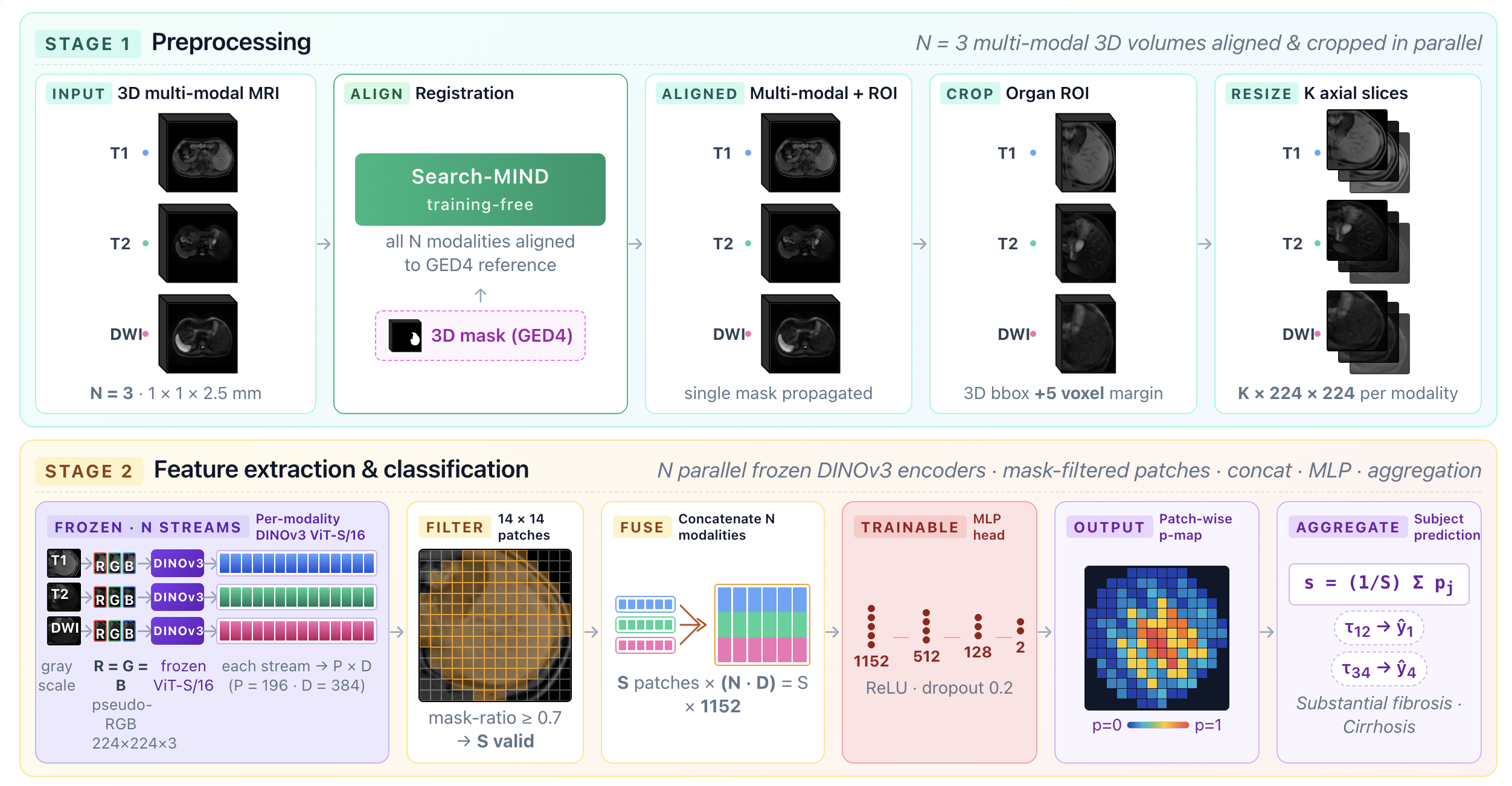}
\caption{Overview of the proposed patch-based multimodal framework. The pipeline begins by co-registering multi-modality 3D volumes and cropped to a mask-defined ROI of $K \times 224 \times 224$. Then, a frozen DINOv3 encoder extracts a $P\times D$ (number of patches $\times$ feature dimension) feature matrix per modality.  $S$ valid patches are selected from the ROI. Features from $N$ modalities are concatenated into an $S \times (N \times D)$ matrix, then passed through an MLP to generate patch-wise probability maps. Finally, an aggregation module pools these results for subject-level classification.} 
\label{fig1}
\end{figure}

The proposed patch-based framework (Fig.~\ref{fig1}) leverages DINOv3 representations for subject-level classification in multimodal medical imaging through a two-stage pipeline. In the preprocessing stage, raw unaligned 3D volumes are co-registered to a common space and cropped to a mask-defined Region of Interest (ROI), yielding standardized $K \times 224 \times 224$ axial slices. During feature representation and classification, a frozen DINOv3 encoder extracts $P \times D$ feature matrices per modality, where $P$ is the total number of patches and $D$ is the embedding dimension. Then, a mask-guided selection module is applied to select $S$ valid patches from the ROI. These features of valid patches are concatenated across the $N$ modalities into a unified $S \times (N \times D)$ matrix, processed by a lightweight Multi-layer Perceptron neural network (MLP) to generate patch-wise probability maps, and finally pooled by an aggregation module for the subject-level prediction. While described generally, this methodology is validated in Section~\ref{sec:experiments} using liver T1, T2, and DWI MRI sequences for fibrosis staging.

\subsection{Preprocessing}

To standardize multimodal 3D volumes for feature extraction, we first employ a registration model to establish spatial correspondence \cite{wang2026searchmindtrainingfreemultimodalmedical}. This alignment allows a reference mask to be propagated across all $N$ modalities, ensuring uniform ROI application without modality-specific segmentations. Once aligned, a 3D bounding box isolates the target organ, expanded by 5 voxels in all directions to retain critical periorgan context. Finally, all cropped ROIs and masks are resized to $224 \times 224$ per axial slice. We preserve the original volume depth ($K$), resulting in a standardized input dimension of $K \times 224 \times 224$ for each modality.

\subsection{Feature Extraction and Patch-level Classification}
\label{sec:MLP}

As a foundation model trained on massive natural image datasets, DINOv3 extracts high-quality features that transfer effectively to downstream tasks without fine-tuning \cite{simeoni2025dinov3}. However, adapting this architecture to medical imaging presents a challenge: DINOv3 requires 3-channel RGB input. Consequently, mapping multimodal MRI sequences into this format is a critical design choice, as the fusion strategy directly influences the quality of the transferred representations.

In our proposed scheme, for each modality, the 2D slice is duplicated across all three RGB channels to form a pseudo RGB image of shape 224$\times$224$\times$3 to feed into DINOv3. Each patch in each modality generates a feature vector with length $D$=384. These vectors of all modalities are then concatenated together to represent the feature of the patch.  

As only the patches within the ROI should be considered for classification modeling, each $224 \times 224$ axial slice is partitioned into $14 \times 14$ ($P$=196) non-overlapping $16 \times 16$ patches. To filter out non-target anatomy and image borders included during bounding box expansion, we retain only patches with a mask-to-patch area ratio exceeding 0.7. This threshold was selected to balance signal purity with data retention. It effectively minimizes non-organ noise while preserving critical boundary patches near the organ capsule that provide essential morphological information. This process results in $S$ valid patches, each represented by an $N \times D $ feature vector. 

Subsequently, a lightweight MLP, consisting of two hidden layers (512, 128), each followed by ReLU and dropout (rate 0.2), is then trained on these features to perform patch-level binary classification. In this scheme, DINOv3 encodes each modality independently. Therefore, self-attention operates only within a modality, and cross-modality fusion is performed by the MLP head via feature concatenation.

\subsection{Patch-to-Subject Aggregation}
Following feature extraction, we aggregate the patch-level predictions to obtain the final subject-level diagnosis using liver fibrosis staging as a case-study. 

In clinical practice, liver fibrosis is categorized into four progressive stages ($S1$–$S4$), ranging from mild fibrosis ($S1$) to cirrhosis ($S4$). Because differentiating between the intermediate stages ($S2$ and $S3$) is notoriously difficult, even for experienced pathologists, direct four-class classification is highly challenging. Following the same evaluation protocol of CARE 2025 challenge \cite{care2025_track4}, the problem is decomposed into binary sub-tasks: Substantial Fibrosis Detection ($S1$ vs. $S2$–$S4$) and Cirrhosis Detection ($S4$ vs. $S1$–$S3$).

The patch-level model is trained as a binary classifier using only the $S1$ (labeled 0) and $S4$ (labeled 1) classes. This design is driven by the hypothesis that fibrotic progression is frequently spatially heterogeneous, manifesting in localized regions rather than uniformly altering the entire liver parenchyma, particularly during the intermediate stages (S2, S3). 

To determine a subject-level diagnosis, we evaluate all valid patches $S$ within the liver ROI. Let $z_j \in \{0, 1\}$ denote the binary prediction for the $j$-th patch. The subject-level score $s$ is then calculated as the proportion of patches identified as Stage 4-like across the entire volume: 
 
\begin{equation}
    s = \frac{1}{S} \sum_{j=1}^{S} z_j
    \label{eq:subject_score}
\end{equation}

To calculate the liver fibrosis score for each subject, we apply a fibrosis probability mapping that translates the raw patch-level proportion $s$ into clinical likelihoods. Recognizing that the decision boundaries for mild fibrosis ($S1$) and cirrhosis ($S4$) are distinct, we utilize two independent, piecewise linear functions anchored by thresholds $\tau_{12}$ and $\tau_{34}$:

\begin{align}
    \hat{y}_1 &= \begin{cases} 
        1 - \frac{0.5}{\tau_{12}}s, & 0 \le s \le \tau_{12} \\ 
        0.5 - \frac{0.5}{1-\tau_{12}}(s - \tau_{12}), & \tau_{12} < s \le 1 
    \end{cases} \\[10pt] 
    \hat{y}_4 &= \begin{cases} 
        \frac{0.5s}{\tau_{34}}, & 0 \le s \le \tau_{34} \\ 
        \frac{0.5(s-\tau_{34})}{1-\tau_{34}} + 0.5, & \tau_{34} < s \le 1 
    \end{cases}
\end{align}
Both thresholds, $\tau_{12}$ and $\tau_{34}$, are optimized on a validation set to account for clinical data distributions. Notably, while the patch-level model is trained exclusively on S1 and S4 images, the thresholds are optimized using the full range of validation data (S1, S2, S3, and S4). We strictly apply this identical scoring and mapping procedure across all evaluated configurations. Consequently, any observed variations in subject-level performance can be attributed exclusively to the efficacy of the chosen feature extraction and fusion mechanisms.

\section{Experiments and Results}
\label{sec:experiments}
\subsection{CARE-Liver 2025 Dataset}
The CARE 2025 Liver Track 4 dataset \cite{liu2025merit,wu2022meru,gao2023reliable} contains multi-parametric MRI scans from 610 patients, acquired across multiple clinical centers using three distinct scanners: the Philips Ingenia (3.0T), Siemens Skyra (3.0T), and Siemens Aera (1.5T). Each subject’s record includes a subset of sequences, including T1-weighted (T1), T2-weighted (T2), Diffusion-Weighted Imaging (DWI), and four Gd-EOB-DTPA-enhanced dynamic phases (GED1–GED4). While other sequences vary by subject, the GED4 phase is available for every subject in the dataset. Because fibrosis stage labels were restricted to the training cohort, we focused our analysis on the 360 subjects with available ground truth. We utilized the non-contrast modalities (T1, T2, and DWI), which comprise 97 cases in Stage 1, 64 in Stage 2, 32 in Stage 3, and 167 in Stage 4.

To address the lack of pre-aligned data and the scarcity of liver segmentation masks (only 30 subjects were annotated), we implemented a preprocessing pipeline. First, all volumes were resampled to a uniform voxel size of $1\text{mm} \times 1\text{mm} \times 2.5\text{mm}$ and cropped/resized to a fixed dimension of $256 \times 256 \times 48$. To generate the missing liver masks, we applied a semi-supervised segmentation method based on the BRBS framework \cite{he2022learning} to the GED4 sequences. Finally, to ensure the GED4-derived masks could be used for other modalities, the T1, T2, and DWI sequences were spatially aligned to the GED4 coordinate space using the Search-MIND registration framework \cite{wang2026searchmindtrainingfreemultimodalmedical}. For any missing modalities, feature vectors of 1s are used for all methods evaluated in this paper. 

\subsection{Baseline Methods and Experimental Settings}
To compare to our proposed method, three baseline methods are implemented. The evaluation protocol, evaluation metrics and parameter settings are also detailed in this section. 

\textbf{Radiomics-Based Classification:}
This baseline follows the classical radiomics paradigm, where expert-designed features are used in place of learned representations. Using PyRadiomics \cite{van2017computational}, we extract 70-dimensional texture features, including GLCM, GLRLM, GLSZM, and GLDM, from the $S$ selected $16 \times 16$ patches across $N$ modalities (T1, T2, and DWI). For each patch, the features from different modalities are concatenated and processed by the same lightweight MLP head described in Section 3.3. This allows us to generate patch-wise probability maps and perform subject-level classification using handcrafted features as a direct comparison to the DINOv3-based approach.

\textbf{Intensity-Based ResNet:} 
As a second baseline, we employ the intensity-based ResNet approach proposed by Wang et al. \cite{wang2026semi}. In this setup, raw intensity values from the three modalities (T1, T2, and DWI) are stacked into a $16 \times 16 \times 3$ tensor and fed into the network for end-to-end patch-level binary classification. We utilize ResNet-18 \cite{he2016deep} as the backbone. Its architecture is specifically suited for the small $16 \times 16$ patch size, as deeper models risk reducing the feature map to a single pixel too early. To ensure a fair comparison with the DINOv3 pipeline, we replace the original ResNet-18 fully connected layer with the same lightweight MLP head described in Section 3.3. Additionally, all training hyperparameters, including the optimizer, learning rate, batch size, and epoch count, are matched exactly to those used for the MLP head in Section~\ref{sec:MLP}. Notably, this ResNet-based network has previously attained one of the best performances on the liver assessment task within the CARE2025 Challenge.

\textbf{SAM-Based Classification:}
SAM-Med2D \cite{cheng2023sam} is an additional baseline based on a large-scale pretrained model to compare with. This domain-specific variant of SAM\cite{kirillov2023segment} provides a robust, segmentation-driven baseline that has been extensively fine-tuned by utilizing comprehensive, multi-modal medical datasets. Although SAM-Med2D is purely engineered for the segmentation task, we adapt its underlying frozen image encoder to serve as a patch-level feature extractor. Due to the input requirements being different from the DINOv3 input size, we resize the input slices to $256\times256$. Each patch produces a feature with a length of 768. To maintain strict experimental parity with DINOv3, we apply the same single-modality replication and feature concatenation protocol to SAM-Med2D as detailed in Section~\ref{sec:MLP}. 

\textbf{Nested Evaluation Protocol:}
Given the limited sample size inherent to clinical cohorts, we adopt a nested cross-validation protocol to ensure unbiased and repeatable performance estimation.

The 360-subject cohort is stratified-sampled into a 90\% development pool (324 subjects) and a 10\% held-out test set (36 subjects). This partition is repeated \emph{10 times with independent random seeds}, yielding 10 independent experimental runs, each with its own test cohort. All subject-level metrics reported in Table~\ref{tab:result_registration_comparison} are the mean $\pm$ standard deviation computed across these 10 runs.

Within each outer run, the 324 development subjects are further partitioned by 4-fold cross-validation into four train/validation splits (roughly 243/81 subjects each). Four patch-level S1/S4 classifiers are trained (one for each fold), and each validation fold is used exclusively to calibrate a pair of fibrosis-probability thresholds $(\tau_{12}, \tau_{34})$ via the piecewise-linear mapping in Eq.~(2)--(3). Due to the unbalanced numbers of S1 and S4, random duplication is used to boost the number of S1 class in the training set. Each of these four models is then tested on the 10\% held-out test set in the outer run. 

In total, the protocol trains $10 \times 4 = 40$ models, and each model is tested on the corresponding 10\% held-out test set. Table~\ref{tab:dataset} shows the quantitative subject distribution across all four fibrosis stages for a single representative outer run.

\begin{table}[t]
    \centering
    \caption{Subject distribution across a single outer run's splits (training / validation / held-out 10\% test).}
    \setlength{\tabcolsep}{10pt} 
    \renewcommand{\arraystretch}{1.2} 
    \begin{tabular}{lccccc}
        \hline
        \textbf{Dataset} & \textbf{S1} & \textbf{S2} & \textbf{S3} & \textbf{S4} \\
        \hline
        Training         & 70 &  0  & 0    &   112    \\
        Validation       & 17    & 58    &  29    &  38  \\
        Test             & 10   &  6  &   3   &  17         \\
        \hline
    \end{tabular}
    \label{tab:dataset}
\end{table}

\textbf{Evaluation Metrics:}
Following the CARE2025 Challenge evaluation protocols, two clinically critical binary subtasks are utilized to evaluate the classification performance for the liver fibrosis staging: Cirrhosis Detection (S1-S3 vs. S4) and Substantial Fibrosis Detection (S1 vs. S2-S4). The Area Under the Receiver Operating Characteristic curve (AUC) and Classification Accuracy (Acc) were computed. Statistical significance (Wilcoxon signed-rank test) is denoted by $^*$ (p < 0.05, DINO vs. Radiomics),† (p < 0.05, DINO vs. ResNet), and $\ddagger$ ($p < 0.05$, DINO vs.\ SAM-Med2D).

\textbf{Parameter settings:} For feature extraction, we utilized the official DINOv3 ViT-Small/16 checkpoint (dinov3-vits16-pretrain-lvd1689m), pretrained on the LVD-1689M dataset comprising 1.69 billion curated natural images \cite{simeoni2025dinov3}. The ViT-S/16 architecture consists of 12 transformer blocks with a 384-dimensional hidden feature space and a native $16 \times 16$ patch size, totaling approximately 21M parameters. The official SAM-Med2D checkpoint sam-med2d\_b was used as the SAM-Med2D baseline. For the downstream binary classification phase, models were trained for 30 epochs with a batch size of 256 using the Adam optimizer. The initial learning rate of $1 \times 10^{-4}$ was managed by a step decay scheduler, which reduced the rate by a factor of 0.7 every 10 epochs.

\subsection{Results}
To evaluate how registration precision and model architecture influence subject-level classification, we systematically compared the radiomics, ResNet, SAM-Med2D and DINOv3 models across two registration settings (rigid and deformable) implemented via the Search-MIND framework \cite{wang2026searchmindtrainingfreemultimodalmedical}. Performance metrics (AUC and ACC) for the S1 vs. S4 binary classification task are summarized in Table \ref{tab:result_registration_comparison}.
\begin{table}[t]
    \centering
    \caption{Subject-level classification performance under rigid and 
    deformable registration across the S4 and S1 subsets. Statistical 
    significance (Wilcoxon signed-rank test) is denoted by $^*$ 
    ($p < 0.05$, DINO vs.\ Radiomics), $^\dagger$ 
    ($p < 0.05$, DINO vs.\ ResNet) and $^\ddagger$ ($p < 0.05$, DINO vs.\ SAM-Med2D)}
    \setlength{\tabcolsep}{8pt}
    \renewcommand{\arraystretch}{1.2}
    \resizebox{\textwidth}{!}{
        \begin{tabular}{llcccccc}
            \toprule
            \textbf{Registration} & \textbf{Model} & \textbf{Mean AUC} 
            & \textbf{Mean ACC} & \textbf{S4 AUC} & \textbf{S4 ACC} 
            & \textbf{S1 AUC} & \textbf{S1 ACC} \\
            \midrule
            \multirow{3}{*}{Rigid} 
            & Radiomics & $0.582\pm 0.072$ & $0.644 \pm 0.022$ 
            & $0.561 \pm 0.082$ & $0.547 \pm 0.033$ 
            & $0.602 \pm 0.080$ & $0.740 \pm 0.032$ \\
            & ResNet & $0.753 \pm 0.044$ & $0.708 \pm 0.030$ 
            & $0.738 \pm 0.045$ & $0.678 \pm 0.035$ 
            & $0.768 \pm 0.072$ & $0.739 \pm 0.037$\\
            & SAM-Med2D & $0.780 \pm 0.052$ & $0.728 \pm 0.038$ 
            & $0.789 \pm 0.055$ & $0.719 \pm 0.044$ 
            & $0.772 \pm 0.071$ & $0.738 \pm 0.045$\\
            & \textbf{DINO} & $\mathbf{0.843 \pm 0.041}$ 
            & $\mathbf{0.766 \pm 0.040}^{*\dagger}$ 
            & $\mathbf{0.848 \pm 0.056}$ 
            & $\mathbf{0.758 \pm 0.057}^{*\dagger}$ 
            & $\mathbf{0.838 \pm 0.063}$ 
            & $\mathbf{0.774 \pm 0.054}$ \\
            \midrule
            \multirow{3}{*}{Deformable} 
            & Radiomics & $0.527 \pm 0.061$ & $0.640 \pm 0.020$ 
            & $0.510 \pm 0.058$ & $0.547 \pm 0.034$ 
            & $0.543 \pm 0.091$ & $0.734 \pm 0.027$ \\
            & ResNet & $0.763 \pm 0.051$ & $0.722 \pm 0.028$ 
            & $0.753 \pm 0.038$ & $0.683 \pm 0.033$ 
            & $0.772 \pm 0.080$ & $0.761 \pm 0.034 $ \\
            & SAM-Med2D & $0.785 \pm 0.049$ & $0.721 \pm 0.039$ 
            & $0.798 \pm 0.058$ & $0.713 \pm 0.049$ 
            & $0.772 \pm 0.059$ & $0.729 \pm 0.050$\\
            & \textbf{DINO} & $\mathbf{0.845 \pm 0.037}$ 
            & $\mathbf{0.765 \pm 0.039}^{*\dagger\ddagger}$ 
            & $\mathbf{0.845 \pm 0.056}$ 
            & $\mathbf{0.746 \pm 0.064}^{*\dagger}$ 
            & $\mathbf{0.845 \pm 0.051}$ 
            & $\mathbf{0.784 \pm 0.053}^{\ddagger}$ \\
            \bottomrule
        \end{tabular}
    }
    \label{tab:result_registration_comparison}
\end{table}

Table \ref{tab:result_registration_comparison} shows the DINOv3 pipeline consistently outperforming all baselines. Under rigid registration, DINOv3 achieves a Mean AUC of 0.843 and Mean ACC of 0.766, significantly exceeding SAM-Med2D (0.780/0.728), ResNet (0.753/0.708), and Radiomics (0.582/0.644). Wilcoxon signed-rank tests confirm DINOv3’s Mean ACC is significantly superior to ResNet and Radiomics across both registration settings (marked $* \dagger$).

\begin{figure}[!p] 
    \centering
    \begin{minipage}{0.48\textwidth}
        \centering
        \includegraphics[width=\textwidth]{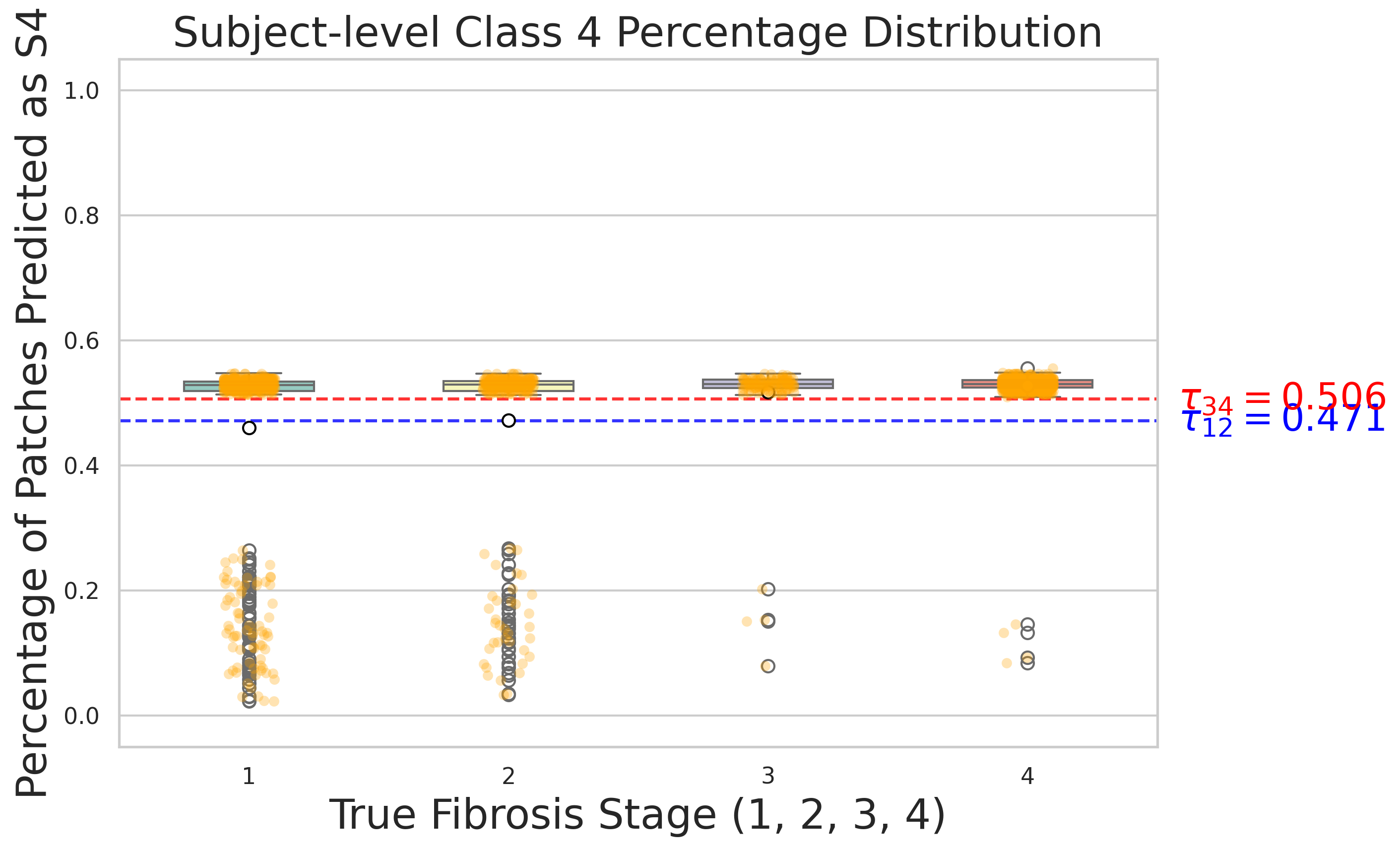}
        \centerline{(a) Radiomics: Rigid}
        \label{fig:m1_rigid}
    \end{minipage}
    \hfill
    \begin{minipage}{0.48\textwidth}
        \centering
        \includegraphics[width=\textwidth]{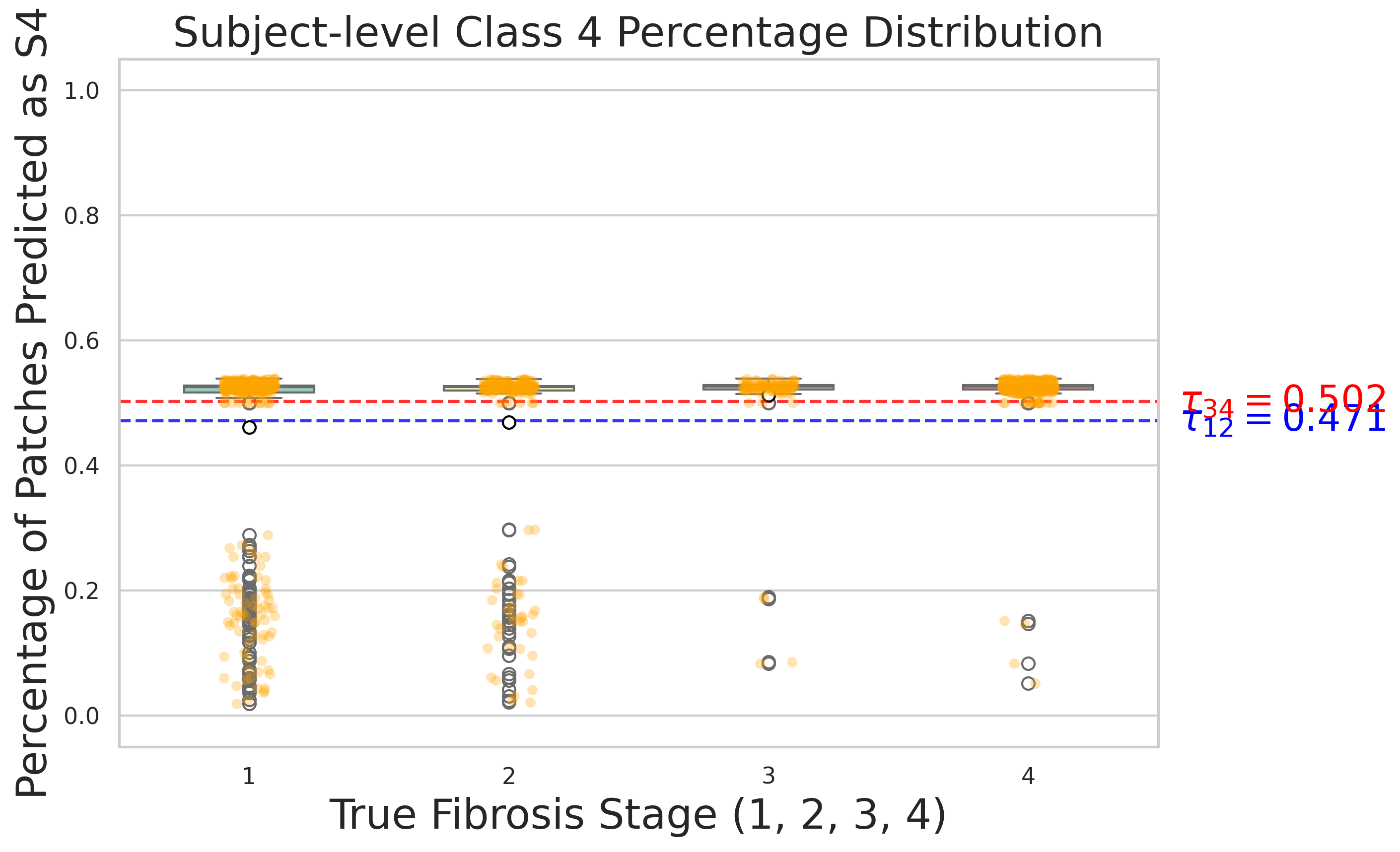}
        \centerline{(b) Radiomics: Deformable}
        \label{fig:m1_deform}
    \end{minipage}

    \vspace{2ex} 

    \begin{minipage}{0.48\textwidth}
        \centering
        \includegraphics[width=\textwidth]{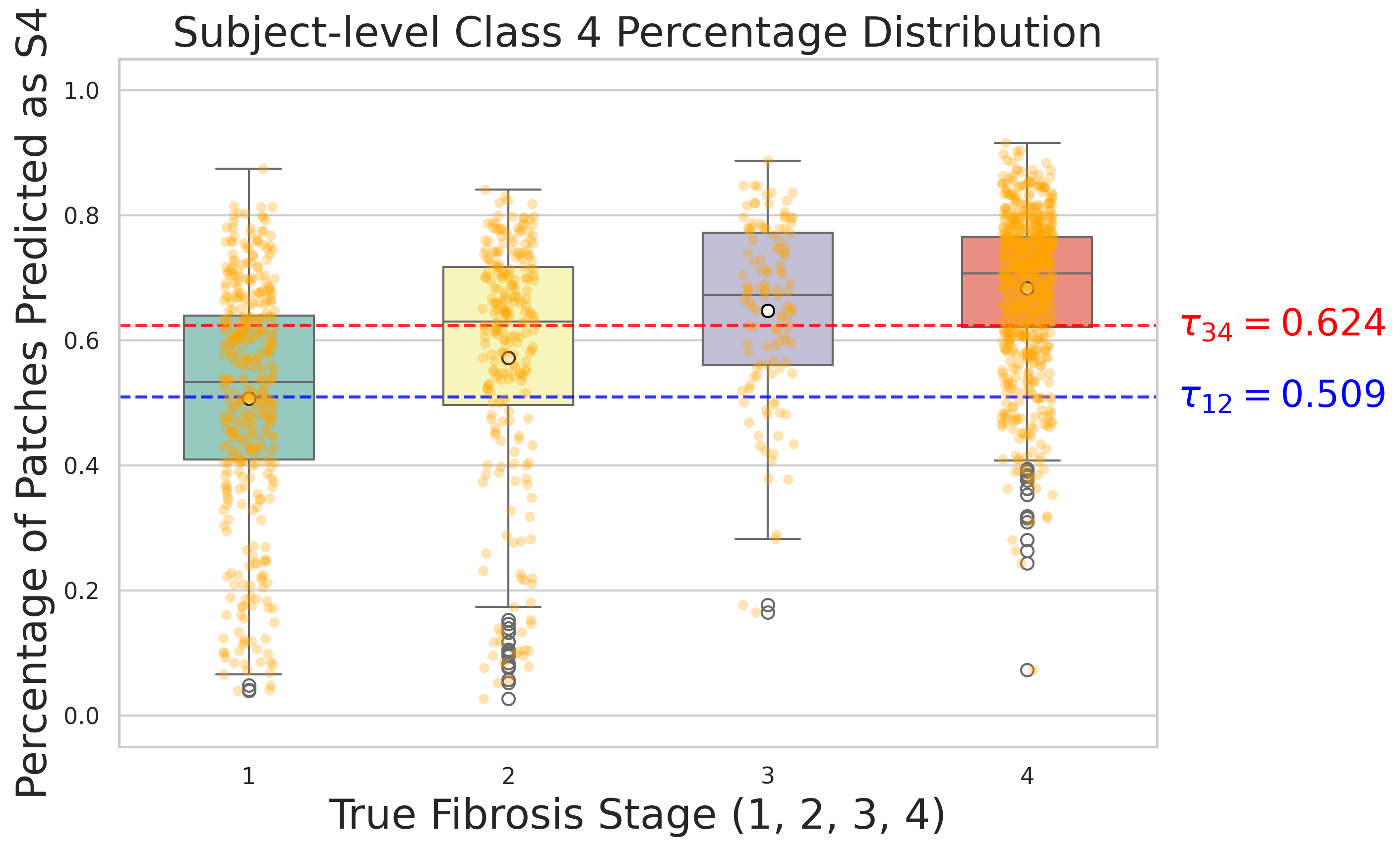}
        \centerline{(c) ResNet: Rigid} 
        \label{fig:m2_rigid}
    \end{minipage}
    \hfill
    \begin{minipage}{0.48\textwidth}
        \centering
        \includegraphics[width=\textwidth]{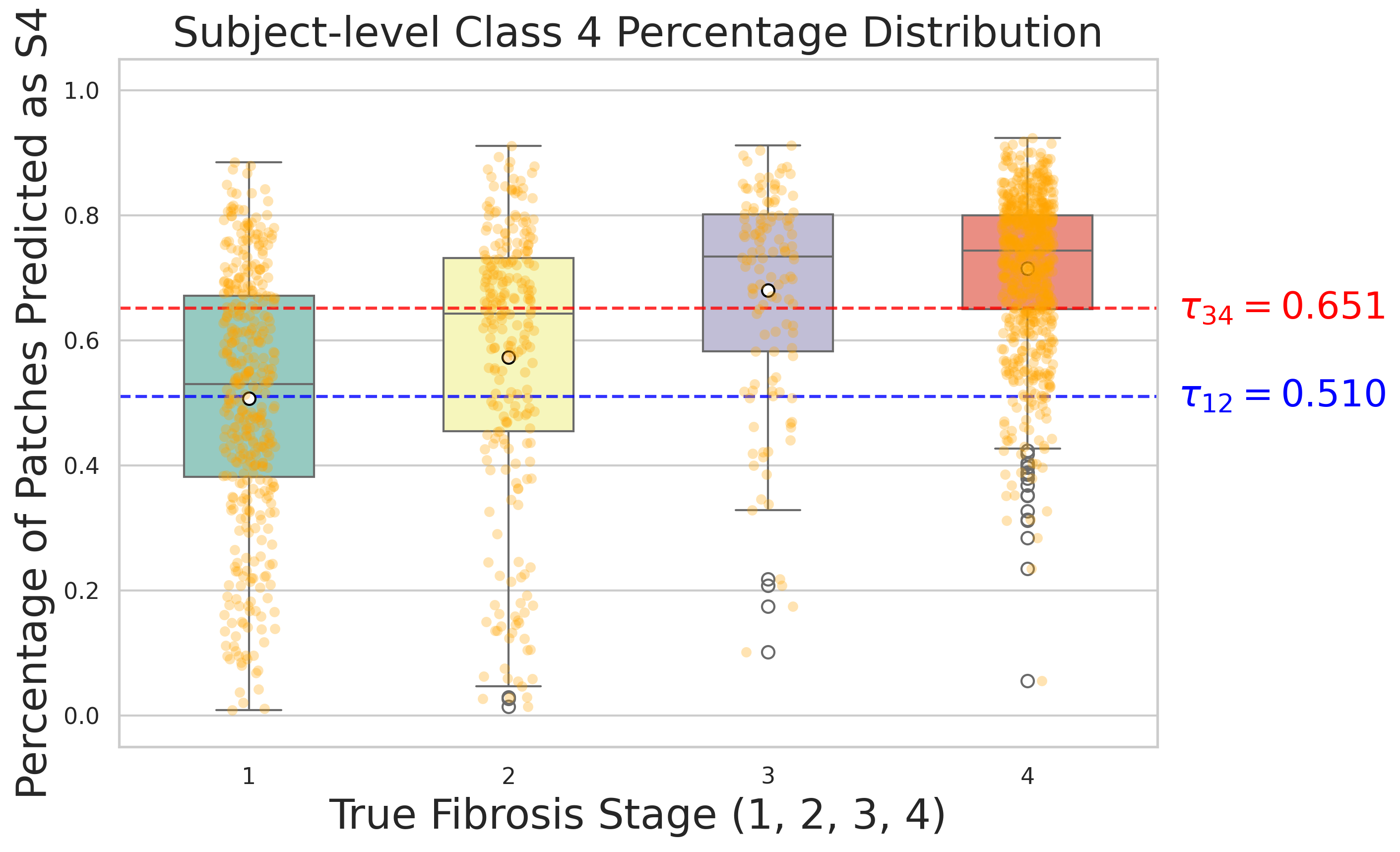}
        \centerline{(d) ResNet: Deformable} 
        \label{fig:m2_deform}
    \end{minipage}
    \vspace{2ex}
    
    \begin{minipage}{0.48\textwidth}
        \centering
        \includegraphics[width=\textwidth]{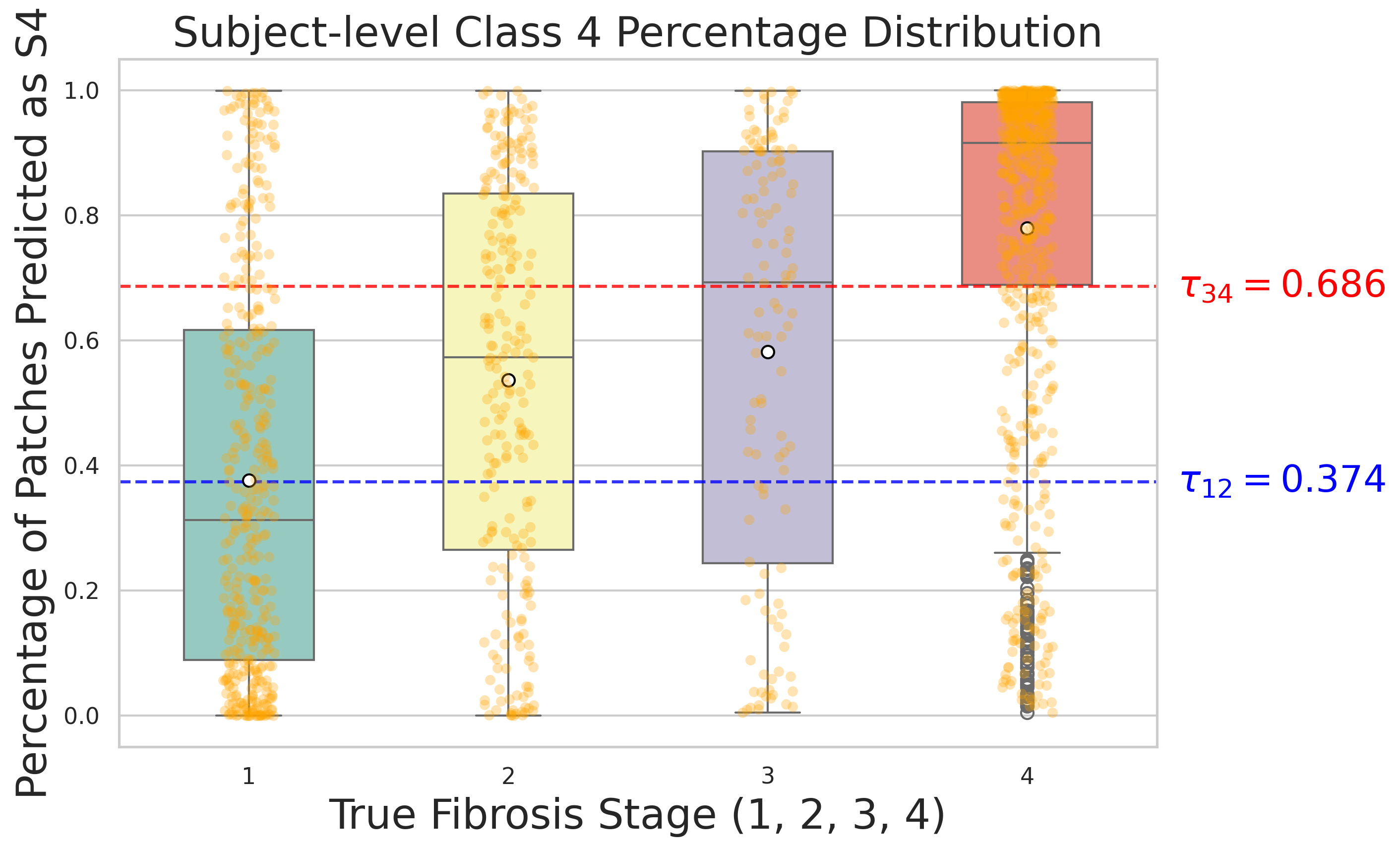}
        \centerline{(e) SAM-Med2D: Rigid} 
        \label{fig:m3_rigid}
    \end{minipage}
    \hfill
    \begin{minipage}{0.48\textwidth}
        \centering
        \includegraphics[width=\textwidth]{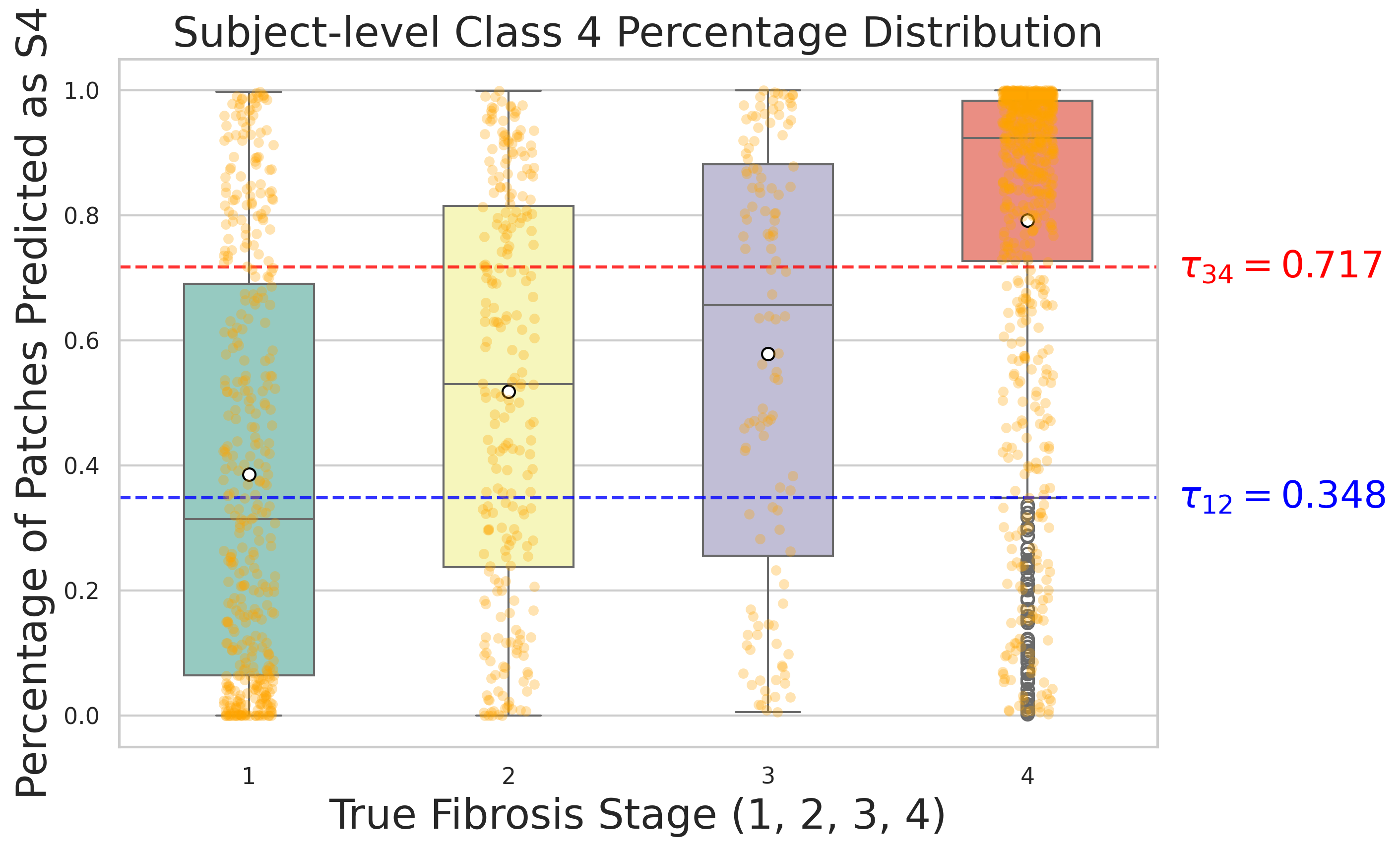}
        \centerline{(f) SAM-Med2D: Deformable} 
        \label{fig:m3_deform}
    \end{minipage}
    \vspace{2ex}

    \begin{minipage}{0.48\textwidth}
        \centering
        \includegraphics[width=\textwidth]{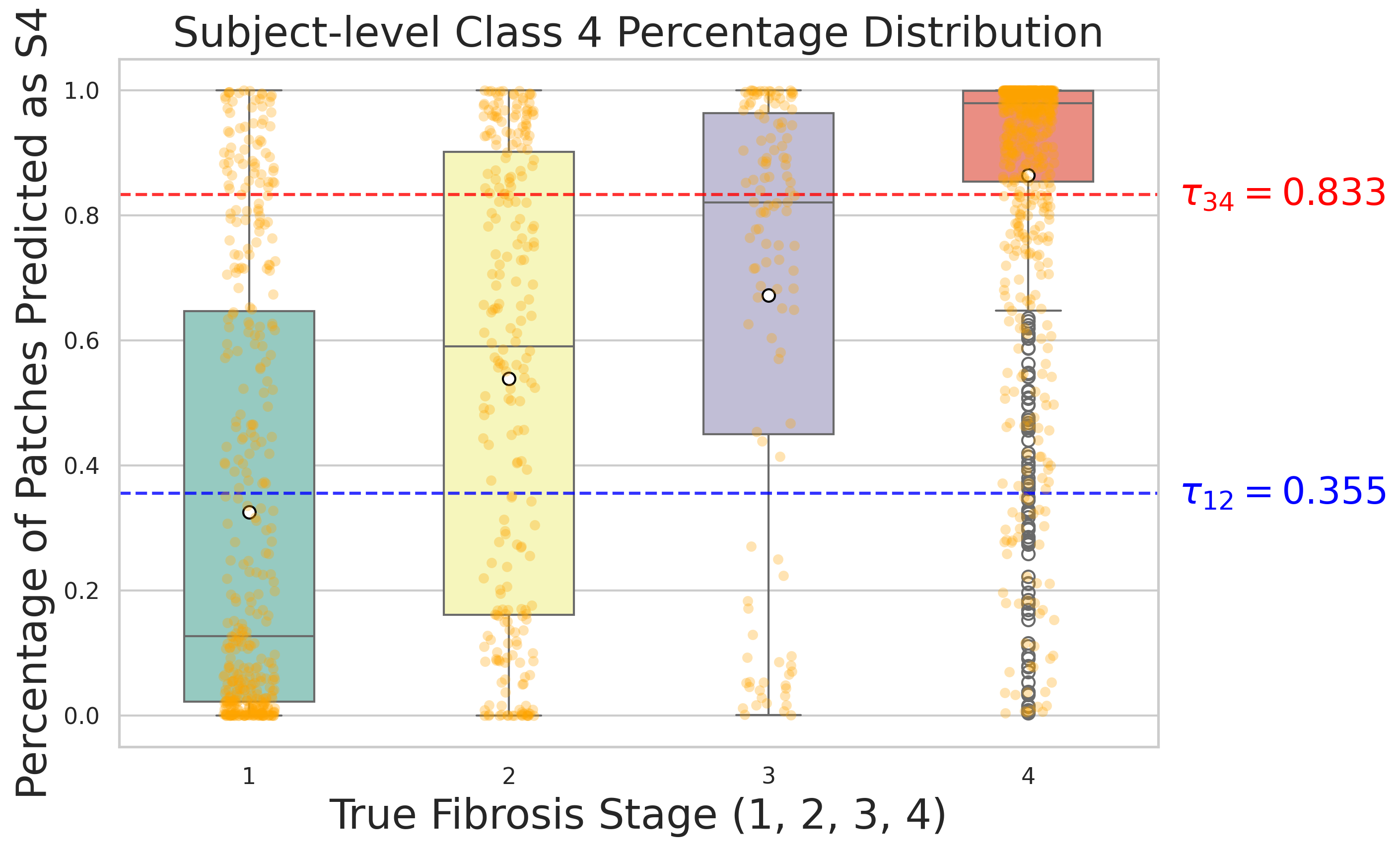}
        \centerline{(g) DINO: Rigid}
        \label{fig:m4_rigid}
    \end{minipage}
    \hfill
    \begin{minipage}{0.48\textwidth}
        \centering
        \includegraphics[width=\textwidth]{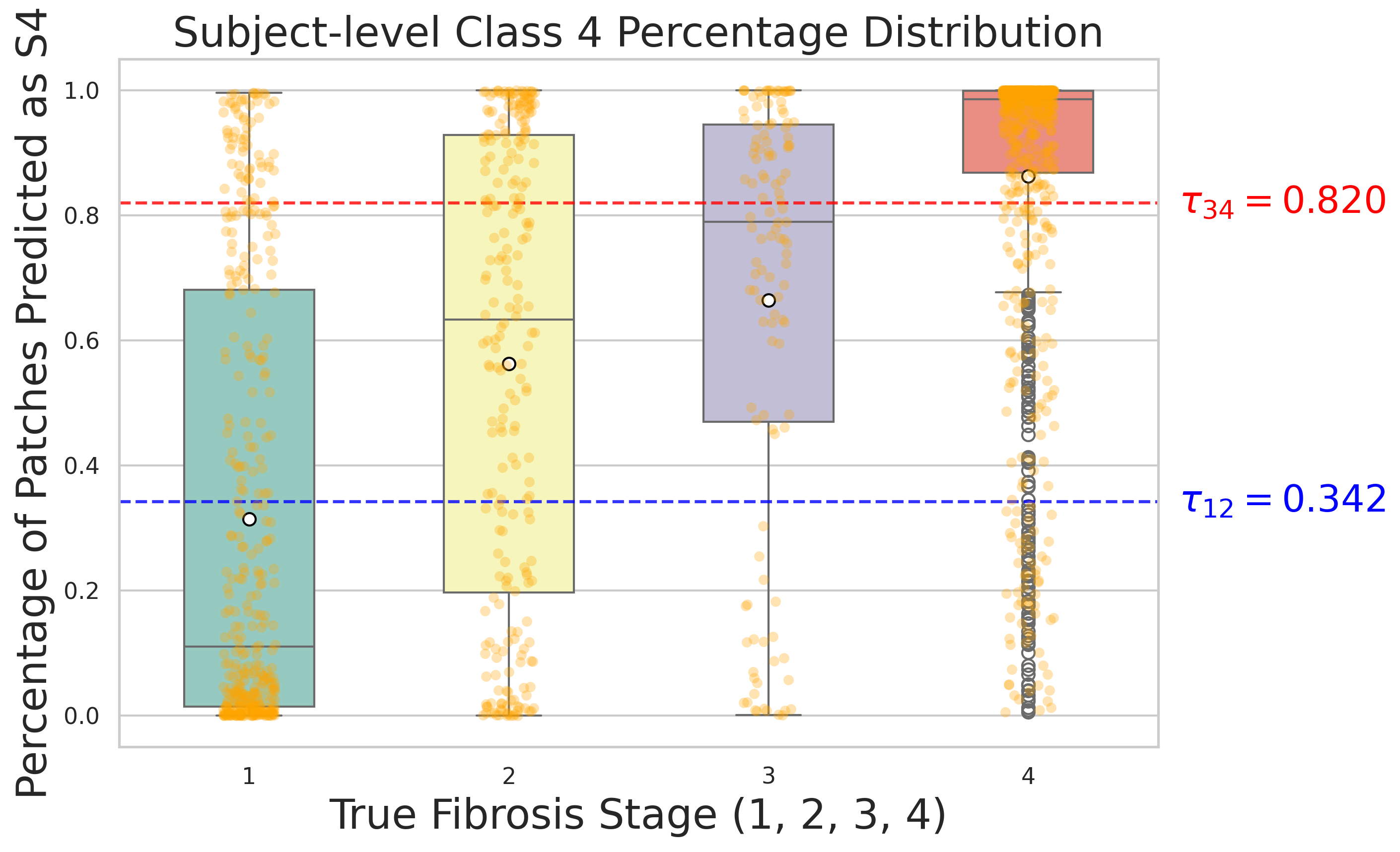}
        \centerline{(h) DINO: Deformable}
        \label{fig:m4_deform}
    \end{minipage}

    \caption{Qualitative comparison of rigid vs. deformable registration across four models. Box plots show the median, interquartile range, and subject-level values (orange dots) over 10 cross-validation rounds. Dashed lines denote thresholds for substantial fibrosis ($\tau_{12}$) and cirrhosis ($\tau_{34}$); the vertical gap between them represents the model’s discriminative margin.}
    \label{fig:total_comparison}
    
\end{figure}

Analysis of the specific stages reveals that the performance gap is most pronounced in the S4 task. DINOv3 achieves approximately 76\% accuracy in both registration settings, outperforming SAM-Med2D by around 4\%, ResNet by roughly 8\% and Radiomics by around 20\%. This suggests that the rich, pre-trained representations of DINOv3 are particularly effective at capturing the macro-morphological changes characteristic of advanced cirrhosis. In contrast, while DINOv3 still maintains the highest scores in the S1 task (reaching 0.784 ACC with deformable registration), the statistical advantage over ResNet and Radiomics model is less significant, likely due to the subtle texture-based nature of early-stage fibrosis. 

The impact of registration strategy (rigid vs. deformable) varied across methods. Radiomics performed better with rigid registration, likely because hand-crafted features are sensitive to the pixel-level interpolations caused by deformable warping. Conversely, the ResNet approach favored deformable registration, as the model could effectively learn from the improved local alignment. Both SAM-Med2D and DINOv3 are pre-trained models, which yielded comparable outcomes regardless of the registration type, indicating greater immunity to local geometric variations.

Figure~\ref{fig:total_comparison} visualizes the distribution of subject-level S4-predicted patch percentages across the four fibrosis stages, incorporating the optimal thresholds $\tau_{12}$ and $\tau_{34}$ from Table \ref{tab:result_threshold_comparison}. This threshold gap serves as a proxy for feature quality: a larger margin indicates superior patch-level representation and class separability. DINOv3 achieves a substantial margin of 0.48 in both registration settings, significantly outperforming other methods. In contrast, radiomics margins fall below 0.05. While ResNet improves on radiomics, its narrow gap suggests difficulty distinguishing intermediate stages (S2, S3) from extremes (S1, S4). SAM-Med2D provides wider margins than ResNet but remains less discriminative than DINOv3.

\begin{table}[!htb]
    \centering
    \caption{Stage 1 and Stage 4 thresholds of patch-based classification under rigid and deformable registration}
    \small 
    \renewcommand{\arraystretch}{1.2}
    \begin{tabular*}{\textwidth}{@{\extracolsep{\fill}}llcc@{}}
        \toprule
        \textbf{Registration} & \textbf{Model} & $\boldsymbol{\tau_{12}}$ & $\boldsymbol{\tau_{34}}$ \\
        \midrule
        \multirow{3}{*}{Rigid} 
        & Radiomics    & $0.471 \pm 0.017$ & $0.506 \pm 0.008$ \\
        & ResNet       & $0.509 \pm 0.048$ & $0.624 \pm 0.035$ \\
        & SAM-Med2D    & $0.374 \pm 0.052$ & $0.686 \pm 0.025$ \\
        & DINO         & $0.355 \pm 0.039$ & $0.833 \pm 0.028$ \\
        \midrule
        \multirow{3}{*}{Deformable} 
        & Radiomics  & $0.471 \pm 0.009$ & $0.502 \pm 0.007$ \\
        & ResNet     & $0.510 \pm 0.037$ & $0.651 \pm 0.037$ \\
        & SAM-Med2D  & $0.348 \pm 0.032$ & $0.717 \pm 0.050$ \\
        & DINO       & $0.342 \pm 0.028$ & $0.820 \pm 0.029$ \\
        \bottomrule
    \end{tabular*}
    \label{tab:result_threshold_comparison}
\end{table}
 
\section{Conclusions}

This study introduces a unified, patch-based framework for liver fibrosis staging that successfully adapts the DINOv3 foundation model to multi-parametric MRI. By integrating training-free registration with hierarchical patch aggregation, our approach consistently outperformed radiomics, ResNet, and SAM-Med2D baselines, achieving peak accuracies of 78.4\% (S1) and 75.8\% (S4). We also observed that large foundation models (DINOv3 and SAM-Med2D) yielded comparable results across both rigid and deformable registration, indicating superior immunity to local geometric variations. Overall, this work demonstrates that frozen vision foundation models provide a powerful, scalable diagnostic baseline without the need for domain-specific fine-tuning. 
Future research will explore lightweight adapters (e.g., LoRA) to bridge the domain gap, extend the framework to cardiac and prostate multi-parametric MRI, and investigate early-fusion and late-fusion strategies to capture cross-sequence spatial dependencies. In addition, our objective is to address the classification of intermediate stages of fibrosis (S2 and S3), a clinically demanding task that typically requires expert consensus and is therefore often omitted from automated staging studies.

%

%
%
\clearpage
\bibliographystyle{splncs04}
\bibliography{refs}

\end{document}